%% file: iclr2021_conference.tex
\documentclass{article}
\usepackage{iclr2021_conference,times}

\input{math_commands.tex}

\usepackage{hyperref}
\usepackage{url}

\usepackage{algorithm}
\usepackage{algorithmic}
\usepackage{stmaryrd}
\usepackage{graphicx}
\usepackage{subfigure}
\usepackage{scalerel}
\usepackage{amssymb}
\usepackage{wrapfig}
\usepackage{csquotes}
\usepackage[bottom]{footmisc}

\newcommand\RCOMMENT[1]{\hfill\(\triangleright\) #1}
\newcommand{\lm}{{\scaleto{\textrm{MLM}}{3pt}}}
\newcommand{\dsr}{{\scaleto{\textrm{DSR}}{3pt}}}
\newcommand{\constraint}{\varobslash}

\title{Distilling Wikipedia mathematical knowledge into neural network models}

\author{Joanne T. Kim\thanks{All authors contributed equally. $^\dagger$Corresponding author.} \\
Lawrence Livermore National Laboratory \\
Livermore, CA 94550, USA \\
\texttt{kim102@llnl.gov} \\
\And
Mikel Landajuela Larma$^*$ \\
Lawrence Livermore National Laboratory \\
Livermore, CA 94550, USA \\
\texttt{landajuelala1@llnl.gov} \\
\AND
Brenden K. Petersen$^{*\dagger}$ \\
Lawrence Livermore National Laboratory \\
Livermore, CA 94550, USA \\
\texttt{bp@llnl.gov} \\
}

\iclrfinalcopy
\begin{document}

\maketitle

\begin{abstract}
Machine learning applications to symbolic mathematics are becoming increasingly popular, yet there lacks a centralized source of real-world symbolic expressions to be used as training data.
In contrast, the field of natural language processing leverages resources like Wikipedia that provide enormous amounts of real-world textual data.
Adopting the philosophy of ``mathematics as language,'' we bridge this gap by introducing a pipeline for distilling mathematical expressions embedded in Wikipedia into symbolic encodings to be used in downstream machine learning tasks.
We demonstrate that a \textit{mathematical language model} trained on this ``corpus'' of expressions can be used as a prior to improve the performance of neural-guided search for the task of symbolic regression.
\end{abstract}

\begin{figure}[b!]
\begin{center}
\includegraphics[trim=0 5 5 0, clip, width=\linewidth]{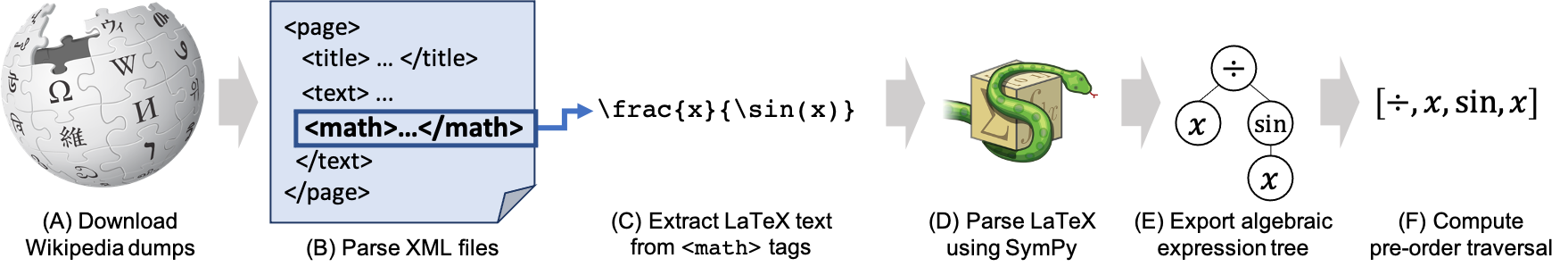} \end{center}
\caption{Pipeline for generating the ``corpus'' of mathematical expressions.}
\label{fig:pipeline}
\end{figure}

\section{Introduction}

\begin{displayquote}
\textit{``The basis of all human culture is language, and mathematics is a special kind of linguistic activity.''}
\vspace{-3mm}
\begin{flushright}
--- \citet{arnold2000mathematics}
\end{flushright}
\end{displayquote}

A growing number of machine learning works leverage datasets of mathematical expressions to perform various symbolic reasoning tasks.
These tasks include symbolic regression \citep{koza1992genetic,petersen2019deep},
symbolic integration \citep{lample2019deep},
solving differential equations, \citep{lample2019deep},
and
freeform mathematical question/answering \citep{saxton2019analysing}.
The expression datasets used to complete these tasks are typically generated procedurally using various rules or heuristics \citep{lample2019deep, saxton2019analysing}, crafted by hand \citep{uy2011semantically}, or manually extracted from existing texts \citep{udrescu2020ai}.
Yet, to the best of our knowledge, there exists no large-scale, customizable, and/or widely used dataset of mathematical expressions.

We introduce a simple pipeline for extracting mathematical expression encodings directly from raw Wikipedia text.
Wikipedia is an excellent source for extracting mathematical knowledge for several reasons.
First, Wikipedia is ever-growing and spans virtually all scientific domains.
Second, expressions embedded in Wikipedia are conveniently annotated via XML tags and follow standard markup encodings (i.e. LaTeX), facilitating extraction and parsing.
Third, Wikipedia is structured into hierarchical categories, allowing users to extract customized expression corpora, e.g. based on a particular branch of science.
Finally, raw Wikipedia data dumps are frequently updated (without requiring web-crawling), easy to access, and freely available.

We demonstrate the practical use of our expression dataset by using it to train a \textit{mathematical language model}, then using the trained model as a prior in a downstream neural-guided search task.
Specifically, we consider symbolic regression, the task of searching the space of tractable mathematical expressions to fit a dataset.
Symbolic regression is an excellent testbed problem for symbolic search because it poses a large combinatorial search space, is computationally inexpensive, and has well-established suites of benchmark problems \citep{white2013better}.
We demonstrate that leveraging a pre-trained mathematical language model as a prior to guide the search improves the ability to recover symbolic expressions from data, and provides other advantages such as reduced semantically invalid expressions.

\section{Related Work}

\textbf{Datasets of expressions.}
Various expression datasets have been proposed for symbolic tasks.
\citet{lample2019deep} procedurally generate expressions to perform symbolic mathematics, specifically to perform integration and solve ordinary differential equations.
Expressions are randomly generated with hand-crafted rules, followed by a cleaning process to simplify and remove invalid expressions.
\citet{saxton2019analysing} released a dataset for mathematical reasoning, which is also generated through rules that sample answer and generates matching questions.
\citet{udrescu2020ai} introduce a dataset of 120 expressions to be used as benchmarks for symbolic regression, manually pulled from the Feynman lectures on physics \citep{feynman1965feynman}.
To the best of our knowledge, there exists no large-scale dataset (or dataset-generating pipeline) of real-world symbolic expressions.

\textbf{Mathematics as language.}
The perspective of viewing mathematics as a form of language dates as far back as the history of mathematics itself \citep{arnold2000mathematics}.
Within machine learning, \citet{lample2019deep} recently addressed symbolic mathematics as a machine translation problem, representing mathematical expressions as sequences and solving mathematical problems with a seq2seq model.
We also consider mathematics as language to create a \textit{mathematical language model} based on human-created, widely-used mathematical expressions in Wikipedia.

\section{Methods}

\begin{wrapfigure}[15]{r}{0.4\linewidth}
\begin{center} 
    \includegraphics[width=0.9\linewidth]{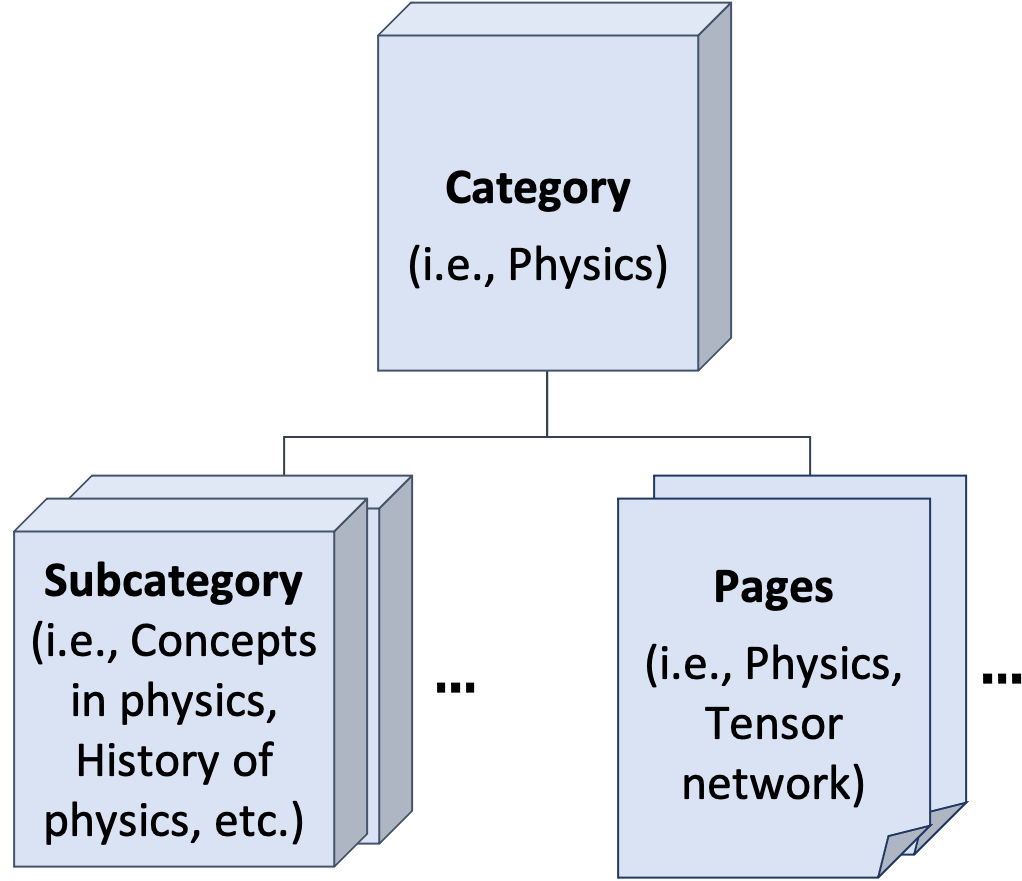}
\end{center}
\caption{Wikipedia hierarchy structure.}
\label{fig:hierarchy}
\end{wrapfigure}

\textbf{Extracting an expression corpus from Wikipedia.}
Our pipeline for extracting expressions from Wikipedia is illustrated in Fig.~\ref{fig:pipeline}.
Wikipedia provides its raw text data as XML (Extensible Markup Language) ``dump'' files.{\footnote{\url{https://dumps.wikimedia.org}}}
Since it is written in a markup language, users can easily parse this dataset of rich information.
Specifically, we extract embedded mathematical expressions, which are annotated via \texttt{<math>} tags (Fig.~\ref{fig:pipeline}b-c).
Raw expressions are represented by LaTeX text, which we convert into a tree-based representation---called an algebraic expression tree---via the computer algebra system SymPy \citep{meurer2017sympy}\footnote{\url{https://sympy.org/}}.
Finally, expressions are encoded as sequences of tokens corresponding to the pre-order traversal of the algebraic expression tree.
This data can then be readily used for downstream machine learning pipelines.

We also leverage Wikipedia's hierarchical page categorizations.
As illustrated in Fig.~\ref{fig:hierarchy}, each category comprises subcategories and pages, forming a tree structure.
Pages are the user interface to access content in Wikipedia, while categories are metadata used to point users to related content.
To build a category tree, we need the edges connecting page and categories.
Edge information is only partially available in the XML dump files.
Therefore, we resort to the database provided by MediaWiki.{\footnote{\url{https://mediawiki.org}}}
In particular, this database includes three relevant tables: Categorylinks, Category, and Page (Fig.~4 in the Appendix).
Using database management system such as MySQL, we use this to generate a category tree from a given root category.
Since categories may not be in strict hierarchy (i.e. they can be recursive), we specify a maximum search depth.
In this manner, users can customize the expression dataset to specific categories (e.g. scientific domain) of interest.

\textbf{Training a mathematical language model.}
By considering each symbolic token as a ``word'' from a dictionary of tokens, and each sequence of tokens (i.e. expression) as a ``sentence,'' we propose building a mathematical language model (MLM) to estimate the probability for a mathematical expression represented by a sequence of tokens.

Before training the MLM, we first filter and augment the dataset based on the operators and operands used in the downstream learning task.
For instance, if the task does not use integrals, each expression containing an integral symbol could be simply removed.
Instead, we choose two approaches to augment this data: replacing and splitting.
When an integral appears while traversing the expressional tree, the whole subtree with a root of integral can be replaced with a terminal token.
Another approach is splitting the tree to use the expression in the integrand for data augmentation.

Finally, we train a recurrent neural network (RNN) with our preprocessed and augmented expression dataset to create a MLM using the set of given symbols as the vocabulary.
Specifically, we train the MLM to predict the next token in a sequence by minimizing the cross-entropy loss between the output of the RNN and the given label.
This strategy is standard in natural language processing (NLP) 
for training language models that predict the next word or character given a partial sequence \citep{mikolov2010recurrent}.

\begin{wrapfigure}[30]{r}{0.5\linewidth}
\begin{center}
\includegraphics[width=1.0\linewidth]{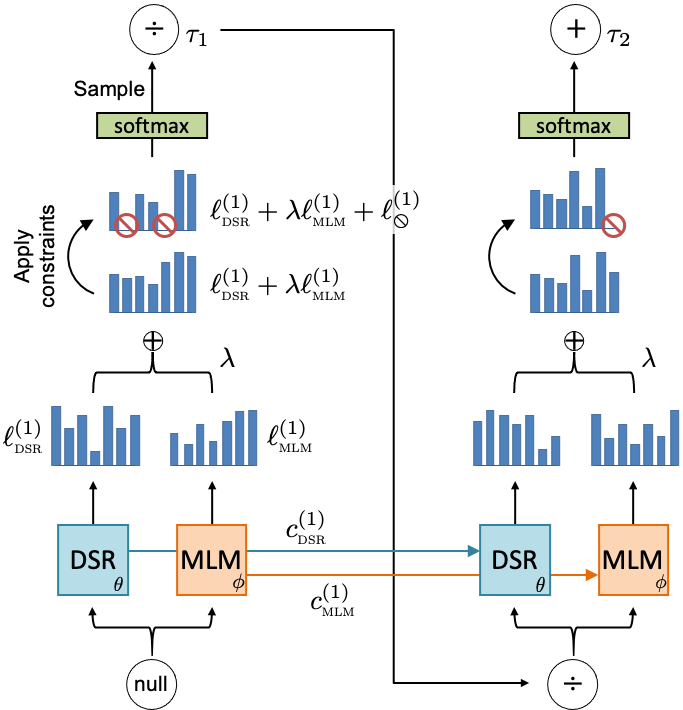}
\end{center}
\caption{Integration between DSR and the MLM prior.
For each token, each architecture (DSR, MLM) outputs a vector of logits.
Logits are combined, constraints are applied, and a final softmax determines the categorical distribution from which a token is sampled.
}
\label{fig:integration}
\end{wrapfigure}

\textbf{Integrating with deep symbolic regression.}
Deep symbolic regression \citep{petersen2019deep} uses an autoregressive recurrent neural network, or \textit{Controller}, parameterized by $\theta$, that emits a categorical distribution on which token to select next, conditioned on all previously selected tokens.

Here, we propose using the MLM to inform the Controller by directly adjusting the logits that are emitted by it, as in Fig.~\ref{fig:integration}.
Specifically, at time step $i$, given input $x$ (e.g. the previously selected token), the Controller emits logits $\ell^{(i)}_\dsr$ and updates its internal state $c^{(i)}_\dsr$:
\begin{align*}
    (\ell^{(i)}_\dsr, c^{(i)}_\dsr) = \textrm{Controller}(x, c^{(i-1)}_\dsr; \theta)
\end{align*}
Further, DSR allows incorporating in situ constraints to the search space (e.g. constraining nested trigonometric operators) via logits $\ell^{(i)}_\constraint$, whose values are either zero or negative infinity (see \citet{petersen2019deep} for details).
To incorporate the MLM into this generative process, we introduce a third vector of logits computed by the MLM:
\begin{align*}
    (\ell^{(i)}_\lm, c^{(i)}_\lm) = \textrm{MLM}(x, c^{(i-1)}_\lm; \phi)
\end{align*}
Finally, the logit vectors are summed, a softmax operator is applied, and the resulting probability vector defines a categorical distribution used to sample the next token $\tau_i$:
\begin{align*}
    \tau_i \sim \textrm{Categorical}(\textrm{Softmax}(\ell^{(i)}_\dsr + \lambda \ell^{(i)}_\lm + \ell^{(i)}_\constraint)).
\end{align*}
The hyperparameter $\lambda \in \mathbb{R}^+$ controls the strength of the MLM prior.
We provide an interpretation of $\lambda$ based on the temperature of the softmax function.
Note the softmax function with temperature $T$ and logits $\ell$ is defined as: $\textrm{softmax}_T(\ell) \circeq \textrm{softmax}(\ell/T)$.
Multiplying logits by $\lambda$ yields:
\begin{align*}
    \lambda \ell = \textrm{inverse-softmax}(\textrm{softmax}(\lambda \ell)) = \textrm{inverse-softmax}(\textrm{softmax}_{\lambda^{-1}}(\ell)),
\end{align*}
where inverse-softmax is defined up to an arbitrary constant.
Thus, $\lambda$ can be viewed as the \textit{inverse temperature} of the MLM prior's contribution to the final softmax.
Higher temperatures (lower $\lambda$) result in a lower influence of the MLM prior on the final softmax.
At the extreme, $\lambda \rightarrow 0$ corresponds to infinite temperature, in which case the MLM prior is ignored.

Pseudocode for DSR integrated with the MLM prior is provided in the Appendix.
Note that the recurrent architecture used in the MLM can be completely different than the one used in DSR; this allows the MLM to be pre-trained offline.

\section{Results and Discussion}

\textbf{Dataset statistics.}
From a raw Wikipedia dump file of $\sim$70 GB, we extracted 798,998 mathematical expressions across 41,763 different pages.
As an example of using the category hierarchies, when we search pages under the category \textit{Physics} with depth 3, we collect 67,404 expressions from 2,265 pages in 879 categories, while \textit{Physics} itself has 25 pages with 1,374 expressions.

\begin{table*}[t]
\centering
\caption{Comparison of recovery rate, mean steps to solve, and mean rate of invalid expressions in the symbolic regression task with and without the MLM prior.}
\begin{tabular}{cccc|ccc}
 & \multicolumn{3}{c}{DSR without MLM}  & \multicolumn{3}{c}{DSR with MLM}  \\
      Benchmark & Recovery & Steps & Invalid       & Recovery & Steps & Invalid  \\ \hline
$x^3+x^2+x$ & 100.\% & 165.2 & 47.82 & 100.\% & 99.47 & 56.75 \\ 
$x^4+x^3+x^2+x$ & 100.\% & 264.6 & 35.86 & 100.\% & 235.6 & 36.46 \\ 
$x^5+x^4+x^3+x^2+x$ & 100.\% & 349.2 & 28.35 & 100.\% & 329.1 & 26.93 \\ 
$x^6+x^5+x^4+x^3+x^2+x$ & 100.\% & 672.2 & 17.65 & 100.\% & 525.5 & 20.75 \\ 
$\sin(x^2)\cos(x)-1$ & 76.\% & 847.8 & 23.67 & 94.\% & 672.8 & 22.81 \\ 
$\sin(x)+\sin(x+x^2)$ & 100.\% & 189.3 & 45.26 & 100.\% & 120.3 & 51.63 \\ 
$\log(x+1)+\log(x^2+1)$ & 35.\% & 1513. & 11.02 & 27.\% & 1620. & 10.91 \\ 
$\sqrt{x}$ & 95.\% & 601. & 31.81 & 99.\% & 365.1 & 32.62 \\ 
$\sin(x)+\sin(y^2)$ & 100.\% & 117.8 & 34.35 & 100.\% & 95.52 & 27.34 \\ 
$2\sin(x)\cos(y)$ & 100.\% & 368.3 & 17.55 & 100.\% & 364.3 & 13.93 \\ 
$x^{y}$ & 100.\% & 22.36 & 56.37 & 100.\% & 12.58 & 41.13 \\ 
$x^4-x^3+\frac{1}{2}y^2-y$ & 0.\% & 2000. & 6.373 & 0.\% & 2000. & 5.71 \\ 
\cline{2-7}
\multicolumn{1}{r}{Average:} & 83.8\% & 592.6 & 29.7\% & \textbf{85.0\%} & \textbf{536.7} & \textbf{28.9\%} 
\end{tabular}
\label{tab:sr-results}
\end{table*}

\textbf{Application to symbolic regression.}
We demonstrate the value of the MLM by using it to inform the task of symbolic regression.
For simplicity, we replicate the experimental setup and hyperparameters detailed in \citet{petersen2019deep}, leveraging the accompanying open-source package ``Deep Symbolic Regression.''
The only change is the introduction of the MLM prior as previously described, sweeping over inverse temperature hyperparameter $\lambda \in \{0.1, 0.2, \dots, 1.0\}$.
The MLM is trained for 200 epochs using a recurrent neural network with a hidden layer size of 256, showing the cross-entropy loss going down to 0.367.

In Table~\ref{tab:sr-results}, we report the recovery rate (fraction of runs in which the exact symbolic expression is found), average number of steps required to find the solution, and the fraction of invalid expressions (those that produce floating-point errors, e.g. overflows) produced during training, over 100 independent runs.
The MLM prior show an improvement in recovery rate, requiring fewer steps and generating fewer invalid expressions.
In particular, we note the dramatic reduction in steps required to find expressions $\sqrt{x}$, $\sin(x)+\sin(y^2)$, and $x^{y}$.
Overall, these results show that the use of the MLM provides a better informed search in symbolic regression tasks.

\section{Conclusion and Future Direction}

We introduce a pipeline for generating a largescale ``corpus'' of mathematical expressions directly from Wikipedia, and demonstrate that a language model trained on this dataset can improve neural-guided search for the task of symbolic regression.
Possible alternative uses of a MLM include auto-completion of writing expressions in LaTeX, or improving recognition of hand-written expressions.

\subsubsection*{Acknowledgments}

This work was performed under the auspices of the U.S. Department of Energy by Lawrence Livermore National Laboratory under contract DE-AC52-07NA27344. Lawrence Livermore National Security, LLC. LLNL-CONF-820039.

\bibliography{iclr2021_conference}
\bibliographystyle{iclr2021_conference}

\clearpage

\appendix

\section{Appendix} \label{sec:appendix}

\begin{figure}[h]
\centering
    \includegraphics[width=0.8\linewidth]{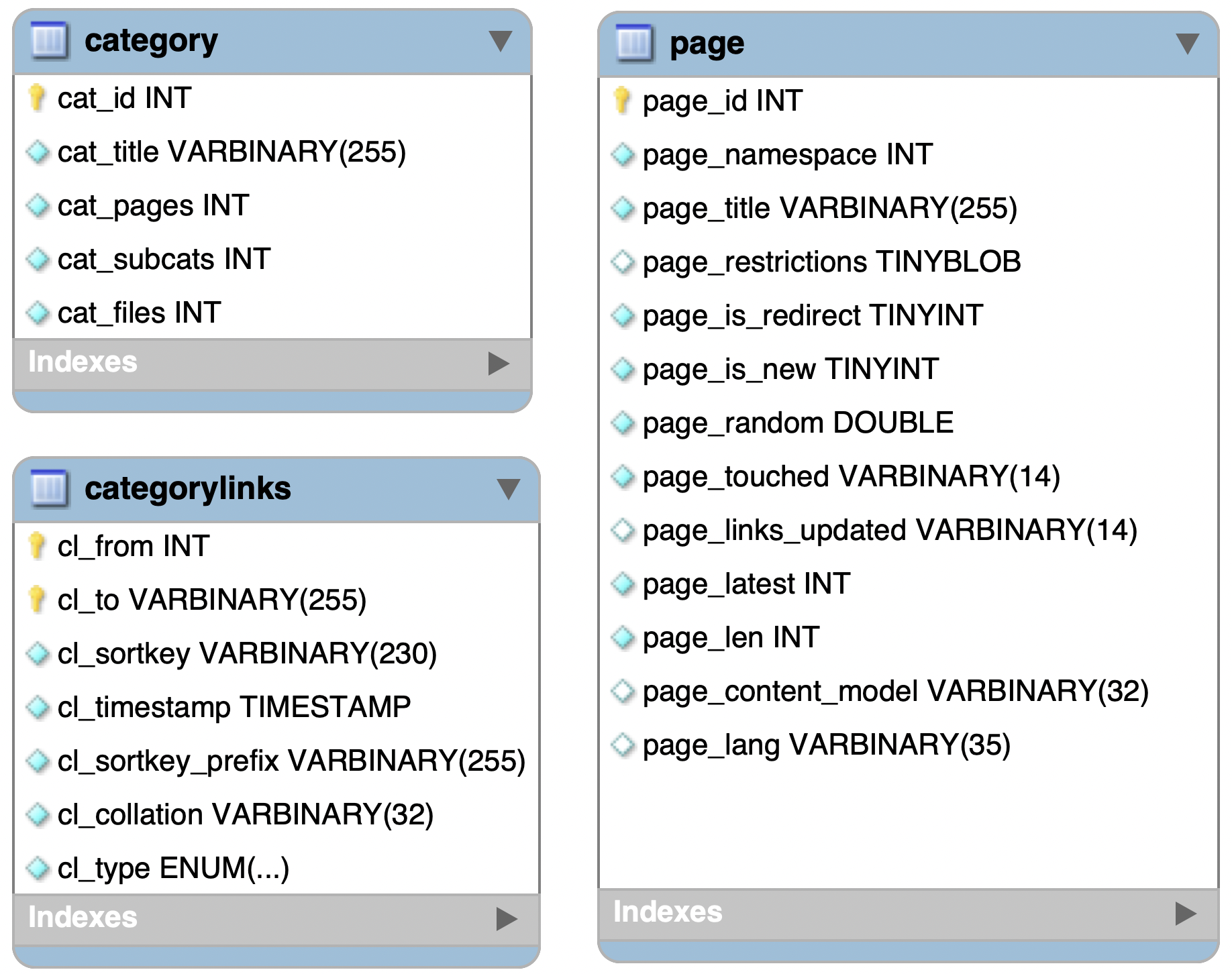}
\caption{Schema of Wikipedia database tables used to extract hierarchy tree of categories and pages. The Categorylinks table holds the subcategory information by setting \texttt{cl\_type} as \textit{subcat}. Note that \texttt{categorylinks.cl\_from} stores the id of the article which matches \texttt{page.page\_id}, a primary key, while \texttt{categorylinks.cl\_to} stores the name of the category as text.}
\label{fig:dbschema}
\end{figure}

\vspace{3em}

\begin{algorithm*}[h]
\caption{Generating an expression from DSR under the MLM}
\label{alg:sampling}
\begin{algorithmic}[1]
\INPUT DSR controller with parameters $\theta$; MLM with parameters $\phi$; library of tokens $\mathcal{L}$
\OUTPUT Pre-order traversal $\tau$ of an expression sampled from the RNN
\STATE $\tau \leftarrow \left[ \right]$ \RCOMMENT{Initialize empty traversal}
\STATE $x \leftarrow \textrm{empty} \| \textrm{empty}$ \RCOMMENT{Initial RNN input is empty parent and sibling}
\STATE $c_0 \leftarrow \vec{0}$ \RCOMMENT{Initialize RNN cell state to zero}
\FOR {$i = 1, 2, \dots$}
\STATE $(\ell^{(i)}_\dsr, c^{(i)}_\dsr) \leftarrow \textrm{Controller}(x, c^{(i-1)}_\dsr; \theta)$ \RCOMMENT{Emit DSR logits; update DSR state}
\STATE $(\ell^{(i)}_\lm, c^{(i)}_\lm) \leftarrow \textrm{MLM}(x, c^{(i-1)}_\lm; \phi)$ \RCOMMENT{Emit MLM logits; update MLM state}
\STATE $\ell^{(i)}_\constraint \leftarrow \textrm{Constraints}(\mathcal{L}, \tau)$ \RCOMMENT{Compute constraints}
\STATE $p^{(i)} \leftarrow \textrm{Softmax}(\ell^{(i)}_\dsr + \lambda \ell^{(i)}_\lm + \ell^{(i)}_\constraint)$ \RCOMMENT{Compute probability vector}
\STATE $\tau_i \leftarrow \textrm{Categorical}(p^{(i)})$ \RCOMMENT{Sample next token}
\STATE $\tau \leftarrow \tau \| \tau_i$ \RCOMMENT{Append token to traversal}
\SHORTIF {$\textrm{expression complete}$}{\textbf{return} $\tau$} \RCOMMENT{If expression is complete, return it}

\STATE $x \leftarrow \textrm{ParentSibling}(\tau)$ \RCOMMENT{Compute next parent and sibling}
\ENDFOR
\end{algorithmic}
\end{algorithm*}

\end{document}

%% file: math_commands.tex
\usepackage{amsmath,amsfonts,bm}

\def\eqref#1{equation~\ref{#1}}

\def\1{\bm{1}}

\DeclareMathAlphabet{\mathsfit}{\encodingdefault}{\sfdefault}{m}{sl}
\SetMathAlphabet{\mathsfit}{bold}{\encodingdefault}{\sfdefault}{bx}{n}













%% file: iclr2021_conference.bbl
\begin{thebibliography}{11}
\providecommand{\natexlab}[1]{#1}
\providecommand{\url}[1]{\texttt{#1}}
\expandafter\ifx\csname urlstyle\endcsname\relax
  \providecommand{\doi}[1]{doi: #1}\else
  \providecommand{\doi}{doi: \begingroup \urlstyle{rm}\Url}\fi

\bibitem[Arnold \& Manin(2000)Arnold and Manin]{arnold2000mathematics}
V~Arnold and Yu~Manin.
\newblock Mathematics as profession and vocation.
\newblock In \emph{Mathematics: Frontiers and Perspectives}, pp.\  153--159.
  American Mathematical Society, 2000.

\bibitem[Feynman et~al.(1965)Feynman, Leighton, Sands, and
  Hafner]{feynman1965feynman}
Richard~P Feynman, Robert~B Leighton, Matthew Sands, and EM~Hafner.
\newblock The feynman lectures on physics; vol. i.
\newblock \emph{American Journal of Physics}, 33\penalty0 (9):\penalty0
  750--752, 1965.

\bibitem[Koza(1992)]{koza1992genetic}
John~R Koza.
\newblock \emph{Genetic Programming: On the Programming of Computers by Means
  of Natural Selection}, volume~1.
\newblock MIT press, 1992.

\bibitem[Lample \& Charton(2019)Lample and Charton]{lample2019deep}
Guillaume Lample and Fran{\c{c}}ois Charton.
\newblock Deep learning for symbolic mathematics.
\newblock \emph{arXiv preprint arXiv:1912.01412}, 2019.

\bibitem[Meurer et~al.(2017)Meurer, Smith, Paprocki, {\v{C}}ert{\'\i}k,
  Kirpichev, Rocklin, Kumar, Ivanov, Moore, Singh, et~al.]{meurer2017sympy}
Aaron Meurer, Christopher~P Smith, Mateusz Paprocki, Ond{\v{r}}ej
  {\v{C}}ert{\'\i}k, Sergey~B Kirpichev, Matthew Rocklin, AMiT Kumar, Sergiu
  Ivanov, Jason~K Moore, Sartaj Singh, et~al.
\newblock Sympy: symbolic computing in python.
\newblock \emph{PeerJ Computer Science}, 3:\penalty0 e103, 2017.

\bibitem[Mikolov et~al.(2010)Mikolov, Karafi{\'a}t, Burget, {\v{C}}ernock{\`y},
  and Khudanpur]{mikolov2010recurrent}
Tom{\'a}{\v{s}} Mikolov, Martin Karafi{\'a}t, Luk{\'a}{\v{s}} Burget, Jan
  {\v{C}}ernock{\`y}, and Sanjeev Khudanpur.
\newblock Recurrent neural network based language model.
\newblock In \emph{Eleventh annual conference of the international speech
  communication association}, 2010.

\bibitem[Petersen et~al.(2021)Petersen, Landajuela, Mundhenk, Santiago, Kim,
  and Kim]{petersen2019deep}
Brenden~K Petersen, Mikel Landajuela, T~Nathan Mundhenk, Claudio~P Santiago,
  Soo~K Kim, and Joanne~T Kim.
\newblock Deep symbolic regression: Recovering mathematical expressions from
  data via risk-seeking policy gradients.
\newblock \emph{Proc. of the International Conference on Learning
  Representations}, 2021.

\bibitem[Saxton et~al.(2019)Saxton, Grefenstette, Hill, and
  Kohli]{saxton2019analysing}
David Saxton, Edward Grefenstette, Felix Hill, and Pushmeet Kohli.
\newblock Analysing mathematical reasoning abilities of neural models.
\newblock \emph{arXiv preprint arXiv:1904.01557}, 2019.

\bibitem[Udrescu \& Tegmark(2020)Udrescu and Tegmark]{udrescu2020ai}
Silviu-Marian Udrescu and Max Tegmark.
\newblock Ai feynman: A physics-inspired method for symbolic regression.
\newblock \emph{Science Advances}, 6\penalty0 (16):\penalty0 eaay2631, 2020.

\bibitem[Uy et~al.(2011)Uy, Hoai, O’Neill, McKay, and
  Galv{\'a}n-L{\'o}pez]{uy2011semantically}
Nguyen~Quang Uy, Nguyen~Xuan Hoai, Michael O’Neill, Robert~I McKay, and Edgar
  Galv{\'a}n-L{\'o}pez.
\newblock Semantically-based crossover in genetic programming: application to
  real-valued symbolic regression.
\newblock \emph{Genetic Programming and Evolvable Machines}, 12\penalty0
  (2):\penalty0 91--119, 2011.

\bibitem[White et~al.(2013)White, Mcdermott, Castelli, Manzoni, Goldman,
  Kronberger, Ja{\'s}kowski, O’Reilly, and Luke]{white2013better}
David~R White, James Mcdermott, Mauro Castelli, Luca Manzoni, Brian~W Goldman,
  Gabriel Kronberger, Wojciech Ja{\'s}kowski, Una-May O’Reilly, and Sean
  Luke.
\newblock Better gp benchmarks: community survey results and proposals.
\newblock \emph{Genetic Programming and Evolvable Machines}, 14\penalty0
  (1):\penalty0 3--29, 2013.

\end{thebibliography}
